%% file: falsesum.tex
\pdfoutput=1

\documentclass[11pt]{article}

\usepackage{falsesum}

\usepackage{times}
\usepackage{latexsym}
\usepackage{graphicx}
\usepackage{booktabs}
\usepackage{multirow}
\usepackage{color,soul}
\usepackage{amssymb}
\usepackage{amsmath}
\usepackage{txfonts}

\usepackage[T1]{fontenc}

\usepackage{microtype}

\usepackage{array}
\newcolumntype{L}[1]{>{\arraybackslash}p{#1}}

\newcommand{\falsesum}{$\textsc{Falsesum}$}
\newcommand{\factcc}{FactCC}
\newcommand\sect[1]{\S\ref{#1}}
\renewcommand{\textsuperscript}[1]{{\footnotesize\raisebox{0.8ex}{#1}}}

\DeclareMathOperator*{\argmax}{max}

\definecolor{cop}{HTML}{3B4D73}
\definecolor{argbg}{RGB}{217, 234, 211}
\definecolor{predbg}{RGB}{208, 224, 227}
\definecolor{codebg}{RGB}{255, 242, 204}
\definecolor{sumbg}{RGB}{239, 239, 239}
\definecolor{blankbg}{RGB}{102, 102, 102}
\definecolor{transp}{RGB}{210, 210, 210}
\definecolor{errorcol}{RGB}{244, 204, 204}

\newcommand{\semitransp}[1]{{\color{transp}\st{#1}}}

\newcommand{\argbox}[1]{\colorbox{argbg}{#1}}
\newcommand{\predbox}[1]{\colorbox{predbg}{#1}}
\newcommand{\codebox}[1]{\colorbox{codebg}{#1}}
\newcommand{\sumbox}[1]{\colorbox{sumbg}{#1}}

\DeclareRobustCommand{\hlpred}[1]{{\sethlcolor{predbg}\hl{#1}}}
\DeclareRobustCommand{\hlarg}[1]{{\sethlcolor{argbg}\hl{#1}}}
\DeclareRobustCommand{\hlcode}[1]{{\sethlcolor{codebg}\hl{#1}}}
\DeclareRobustCommand{\hlsum}[1]{{\sethlcolor{sumbg}\hl{#1}}}
\DeclareRobustCommand{\hlerror}[1]{{\sethlcolor{errorcol}\hl{#1}}}
\DeclareRobustCommand{\hlblank}[1]{{\sethlcolor{blankbg}\color{white}\hl{$\texttt{<span\char`_#1>}$}}}

\title{\falsesum: Generating Document-level NLI Examples \\for Recognizing Factual Inconsistency in Summarization}

\author{Prasetya Ajie Utama\textsuperscript{$\dagger\diamondsuit$} \quad Joshua Bambrick\textsuperscript{$\dagger$} \quad Nafise Sadat Moosavi\textsuperscript{$\ddag\diamondsuit$} \quad Iryna Gurevych\textsuperscript{$\diamondsuit$} \\ \\
\textsuperscript{$\dagger$} Bloomberg, London, United Kingdom \\
\textsuperscript{$\diamondsuit$} UKP Lab, Technical University of Darmstadt, Germany \\
\textsuperscript{$\ddag$} Department of Computer Science, The University of Sheffield \\
\texttt{\{putama,jbambrick7\}@bloomberg.net}
}

\begin{document}
\maketitle
\begin{abstract}
Neural abstractive summarization models are prone to generate summaries which are factually inconsistent with their source documents. Previous work has introduced the task of recognizing such factual inconsistency as a downstream application of natural language inference (NLI). However, state-of-the-art NLI models perform poorly in this context due to their inability to generalize to the target task. In this work, we show that NLI models can be effective for this task when the training data is augmented with high-quality task-oriented examples. We introduce \falsesum{}, a data generation pipeline leveraging a controllable text generation model to perturb human-annotated summaries, introducing varying types of factual inconsistencies. Unlike previously introduced document-level NLI datasets, our generated dataset contains examples that are diverse and inconsistent yet plausible. We show that models trained on a \falsesum{}-augmented NLI dataset improve the state-of-the-art performance across four benchmarks for detecting factual inconsistency in summarization.\footnote{The code to obtain the dataset is available online at \url{https://github.com/joshbambrick/Falsesum}}

\end{abstract}

\section{Introduction}
Recent advances in conditional text generation and the availability of large-scale datasets have given rise to models which generate highly fluent abstractive summaries~\cite{lewis-etal-2019-bart, zhang-etal-2019-pegasus}. However, studies indicate that such models are susceptible to generating factually inconsistent outputs, i.e., where the content of the summary is not semantically entailed by the input document~\cite{kryscinski-etal-2019-neural, goodrich-etal-2019-assessing}. This motivates a new line of research for recognizing factual inconsistency in generated summaries~\cite{kryscinski-etal-2020-evaluating, pagnoni-etal-2021-understanding, wang-etal-2020-asking, fabbri-etal-2021-summeval}.

This factual consistency problem is closely related to the task of natural language inference (NLI) whereby a \textbf{hypothesis} sentence is classified as either entailed, neutral, or contradicted by a given \textbf{premise} sentence~\cite{condoravdi-etal-2003-entailment, dagan-etal-2006-pascal, bowman-etal-2015-large}. Using an input document as the premise and a corresponding generated summary as the hypothesis, earlier solutions have adopted out-of-the-box NLI models to detect factual inconsistency, albeit with limited success~\cite{falke-etal-2019-ranking, kryscinski-etal-2020-evaluating}. 

\begin{figure*}
    \centering
    \includegraphics[width=\linewidth]{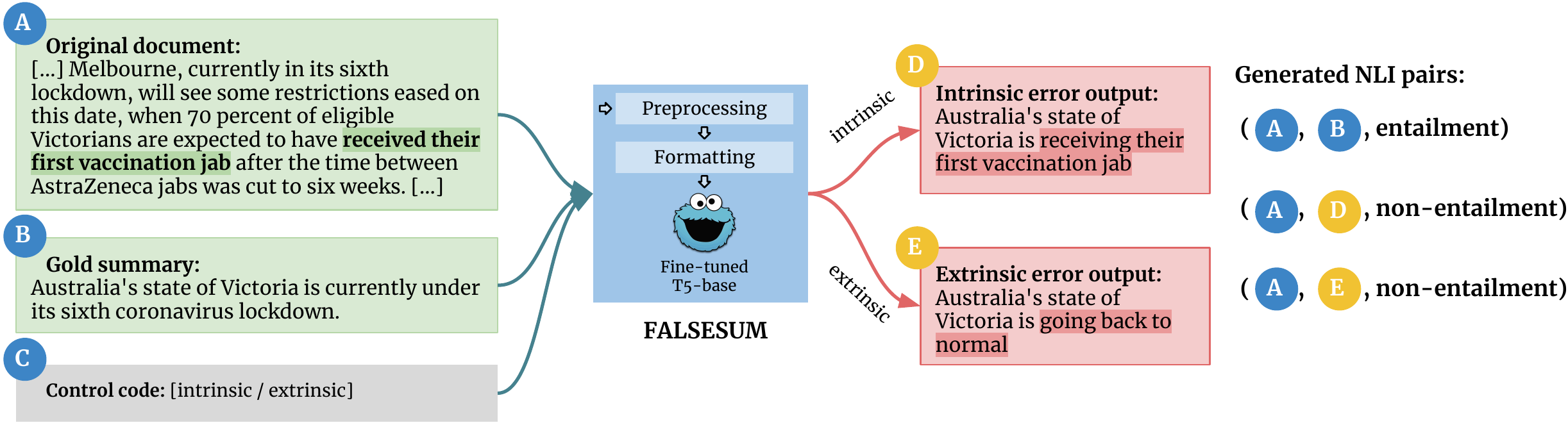}
    \caption{Overview of the \falsesum{} generation framework. \falsesum{} preprocesses and formats the source document ($\mathrm{A}$) and a gold summary ($ \mathrm{B}$) before feeding it to a fine-tuned generator model. The model produces a factually inconsistent summary, which can then be used to obtain $(\mathrm{A}, \mathrm{D})$ \textbf{or} $(\mathrm{A}, \mathrm{E})$ as the negative (non-entailment) NLI premise-hypothesis example pair. We also use the original $(\mathrm{A}, \mathrm{B})$ as a positive NLI example (entailment).}
\label{fig:falsesum_overview}
\end{figure*}

This poor performance largely stems from the fact that most NLI datasets are not designed to reflect the input characteristics of downstream tasks~\cite{khot-etal-2018-scitail}. Such datasets may not always capture the kinds of entailment phenomena which naturally arise from neural abstractive summarization. More importantly, there is also a discrepancy in terms of the input granularity, i.e., the premises in this consistency classification task consist of multi-sentence documents while common NLI datasets use single-sentence premises.

In this work, we introduce \falsesum{}, a data generation pipeline that produces NLI examples consisting of documents paired with gold summaries as \textbf{positive} examples and automatically generated inconsistent summaries as \textbf{negative} examples. We propose a novel strategy to train a text generation model to render false summaries of a given document using only supervision from an existing summarization dataset~\cite{nallapati-etal-2016-abstractive-2}. In addition, our generator supports switchable input control codes to determine the type of factual error exhibited in the generated output. This design allows \falsesum{} to compose diverse and naturalistic outputs which more closely resemble the inconsistent summaries generated by summarization models~\cite{maynez-etal-2020-faithfulness}.
This contrasts with previous solutions~\citep[e.g.,][]{kryscinski-etal-2020-evaluating, yin-etal-2021-docnli}, which synthesize NLI examples using rule-based transformations or language model-based replacements, limiting their diversity and ability to reflect realistic factual errors in summarization. Overall, our contributions in this paper are the following:\\

First, we present a novel training pipeline to create a text generation model which takes as input a pair of a document and a corresponding gold summary. It then perturbs the summary such that it is no longer factually consistent with the original document. Our strategy obviates the need for explicit examples of inconsistent summaries, using only an existing summarization dataset. We use this model to generate a large-scale NLI dataset for the task of recognizing factually inconsistent summaries. The resultant dataset consists of pairs with documents as the premise and naturalistic summaries as the hypotheses, each labeled as either \textbf{entailment} or \textbf{non-entailment}.\\

Second, we demonstrate the utility of our generated data for augmenting existing NLI datasets. We show that on four benchmark datasets, NLI models trained on \falsesum{}-augmented data outperform those trained on previous document-level NLI datasets. We conduct an analysis to show that \falsesum{}-generated summaries are plausible and hard to distinguish from human-written summaries. Lastly, we show that the improvement over the benchmarks is largely attributable to the diversity of factual errors that \falsesum{} introduces.

\section{Related Work}
This work is related to the growing body of research into factual consistency and hallucination in text generation models, particularly for summarization~\cite{cao-etal-2018-faithful}. Research has found that around 30\% of summaries generated by abstractive summarization models contain information which is inconsistent with the source document~\cite{kryscinski-etal-2019-neural}. This motivates the development of an automatic approach to assess factual consistency in generated summaries, in addition to the benchmark datasets to measure the progress in this task~\cite{falke-etal-2019-ranking, kryscinski-etal-2020-evaluating, pagnoni-etal-2021-understanding, fabbri-etal-2021-summeval}. 

Earlier work by~\citet{goodrich-etal-2019-assessing} proposes to use an information extraction model to extract relation tuples from the ground-truth summary text and the generated summary and then count the overlap as the measure of factuality. \citet{eyal-etal-2019-question, durmus-etal-2020-feqa, wang-etal-2020-asking} use a question-answering model to detect factual inconsistency by matching the predicted answers using the document and the summary as the context. 

Concurrently, researchers have drawn a connection between factual consistency and natural language inference (NLI), observing that all information in a summary should be \textbf{entailed} by the source document. While this approach enables the summary to be directly evaluated without first extracting its intermediate semantic structure, earlier attempts were largely unsuccessful. \citet{falke-etal-2019-ranking} use the probabilities assigned to the entailment label by NLI models to re-rank the summary candidates given by beam search but found no improvement in the consistency errors. \citet{kryscinski-etal-2020-evaluating} evaluate out-of-the-box NLI models on the task of inconsistency detection in a binary classification setting and show that the performance is only slightly better than majority voting. 

In the same paper, \citet{kryscinski-etal-2020-evaluating} propose \factcc{}, a synthetic NLI data generation process which applies a set of transformation rules to obtain examples of inconsistent summaries (e.g., sentence negation, entity swapping). They demonstrate that the resulting NLI model performs well on realistic test cases which are obtained by manually annotating the output of several summarization models. This highlights the importance of NLI examples beyond sentence-level granularity and which more closely resemble the input characteristics of the downstream tasks~\cite{mishra-etal-2021-looking}.\footnote{Contemporaneous work by \citet{laban-etal-2022-revisiting} attempts to improve the application of sentence-level NLI models to detect document-level factual inconsistencies using a learnable aggregation of sentence-level predictions. Our work is orthogonal since they can benefit from better quality training examples to train their aggregation weights.}

While the \factcc{} model is moderately effective for detecting factual inconsistency, subsequent work indicates that it only performs well on easier test cases, where highly extractive summaries (i.e., those with high lexical overlap between a summary and the source document) tend to be factually consistent and more abstractive summaries are likely to be inconsistent~\cite{zhang-etal-2020-close}. Furthermore, \citet{goyal-durrett-2021-annotating} show that the synthetic and rule-based nature of \factcc{} leads to lack of diversity of consistency error types and it poorly aligns with the error distribution found in more abstractive summaries. 

\falsesum{} addresses these limitations using controlled natural language generation to construct an NLI dataset which better targets the summarization domain. Inspired by the recent work on controllable generation~\cite{keskar-etal-2019-ctrl, ross-etal-2021-tailor}, we employ a generation model conditioned on an input code which controls the type of consistency errors induced. We further use the generated document-level NLI examples for augmentation and show that NLI models can benefit from the additional data without hurting their existing inference ability~\cite{min-etal-2020-syntactic}. 

\section{\falsesum{} Approach}

\subsection{Design Overview}
\label{sec:design-overview}

\falsesum{} takes as an input a source document $\mathrm{D}$ and a corresponding reference summary $\mathrm{S}^+$. The framework then \textbf{preprocesses} and \textbf{formats} $\mathrm{D}$ and $\mathrm{S}^+$ and feeds them into a generation model $\mathcal{G}$ which outputs a factually inconsistent summary $\mathrm{S}^-$. For each summarization example, we then have both positive \textbf{(entailment)} and negative \textbf{(non-entailment)} NLI tuples $(\mathrm{D}, \mathrm{S}^+, Y=1)$, $(\mathrm{D}, \mathrm{S}^-, Y=0)$, which consist of a document-level premise, a summary sentence, and the consistency label ($1$ indicates entailment). 

\falsesum{} aims to produce a naturalistic $\mathrm{S}^-$ which is contrastive with respect to its corresponding $\mathrm{S}^+$. This means that $\mathrm{S}^+$ and $\mathrm{S}^-$ should be indistinguishable in their surface characteristics (e.g., style, length, vocabularies) and only differ in their factual consistency with respect to $\mathrm{D}$. This ensures that the resulting NLI model learns the correct notion of factual consistency rather than discriminating based on surface features~\cite{mccoy-etal-2019-right}. In addition to naturalness, we consider the diversity of the consistency error types exhibited by $\mathrm{S}^-$. We follow the \textbf{consistency error typology} introduced by~\citet{maynez-etal-2020-faithfulness}, which categorizes consistency errors as either \textbf{intrinsic}, i.e., errors due to incorrect consolidation of information from the source document, or \textbf{extrinsic}, i.e., errors due to assuming \textit{new} information not directly inferable from the contents of the source document.

As illustrated in Figure~\ref{fig:falsesum_overview}, a generation model $\mathcal{G}$ is trained to imitate the consistency mistakes of summarization models. Specifically, it generates perturbed summaries by either \textbf{(1)} incorrectly inserting pieces of information from the source document into random spans of the original summary; or \textbf{(2)} amending pieces of information in the summary by hallucinating new ``facts'' not present in the source document.

To this end, the framework identifies \textit{\textbf{($\diamondsuit$i)}} \textbf{what} information or ``facts'' in the source document are available to the generator; and \textit{\textbf{($\diamondsuit$ii)}} \textbf{where} the incorrect information can be inserted into the gold summary, which is indicated by span \textbf{masking}. We obtain both by subsequently performing \textbf{input preprocessing} and \textbf{formatting} steps (\sect{sec:input-preprocessing} and \sect{sec:input-formatting}).

Next, we define the following seq2seq task to train the model $\mathcal{G}$:
``Given \textit{\textbf{($\diamondsuit$i)}} a list of \textbf{shuffled} and \textbf{formatted} pieces of information extracted from source document and gold summary and \textit{\textbf{($\diamondsuit$ii)}} a partially \textbf{masked} gold summary, fill in the blanks and generate the original gold summary.'' Note that using gold summaries means that we can apply the existing summarization corpus to train $\mathcal{G}$ to generate more coherent and plausible sentences.

\subsection{Input Preprocessing}
\label{sec:input-preprocessing}

Following~\citet{goodrich-etal-2019-assessing}, ``facts'' in the source document and the gold summary are defined as an open information extraction (OpenIE) tuple, which represents the predicate and argument structures found in a sentence. We denote each relation tuple as $( \textsc{arg}_0, \textsc{pred}, \dots, \textsc{arg}_n$), where predicate $\textsc{pred}$ describes the event (\textbf{what} happened) and its complementing semantic arguments $\textsc{arg}$ represent the \textbf{who}, \textbf{to whom}, \textbf{where}, or \textbf{how} of the event. Predicates are usually the main verb of a clause. Both predicates and their arguments consist of spans of tokens~\cite{fader-etal-2011-identifying}. 

\begin{figure}
    \centering
    \includegraphics[width=\linewidth]{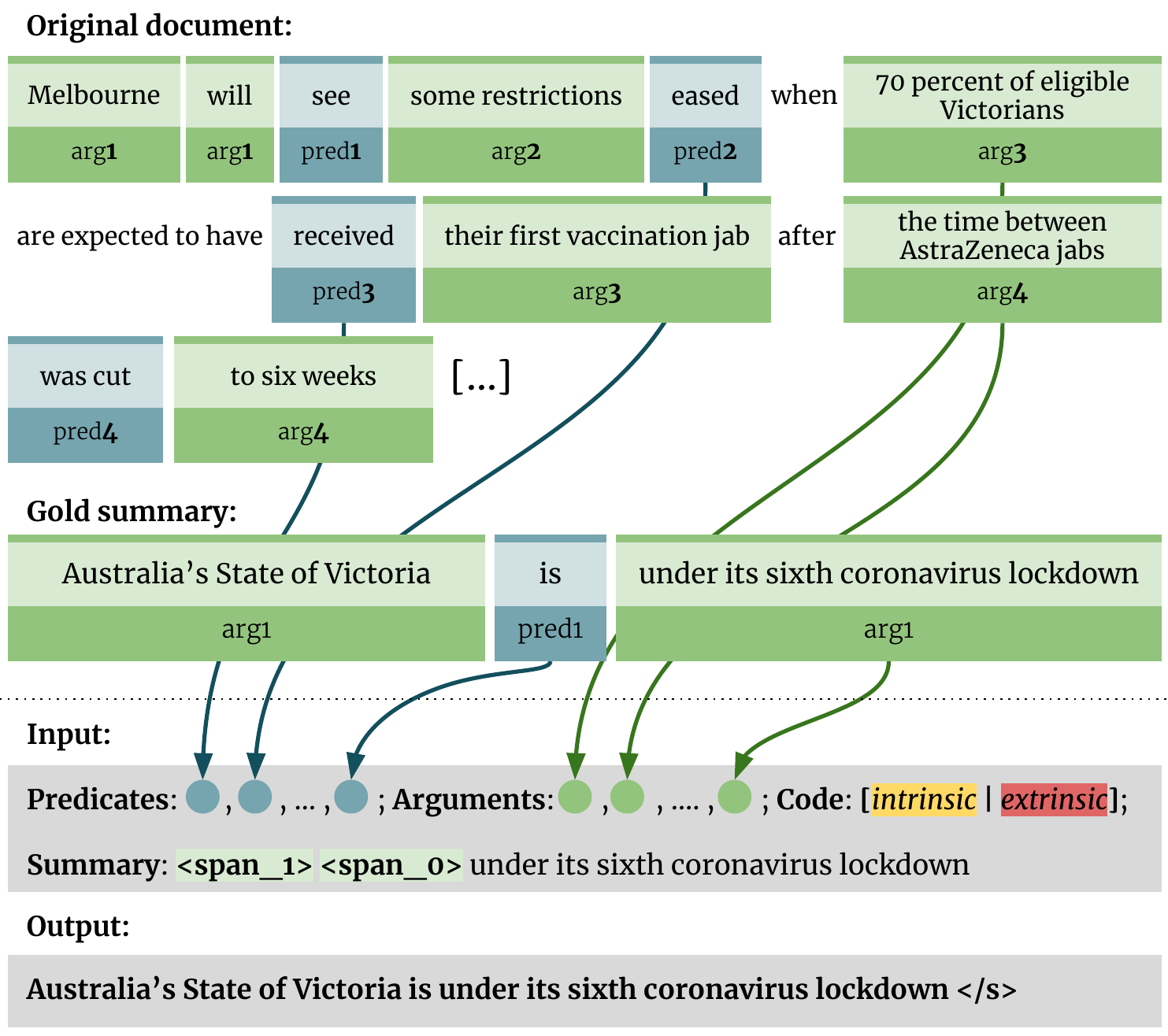}
    \caption{Input format design of \falsesum{}. The framework first extracts the predicate and argument spans from the source document and the gold summary. The spans are then corrupted, lemmatized, and shuffled before being inserted into the input template.}
\label{fig:falsesum_input}
\end{figure}

We use an OpenIE implementation of PredPatt~\cite{white-etal-2016-universal, zhang-etal-2017-evaluation}, a pattern-based framework for predicate-arguments extraction.\footnote{We note that the quality of the OpenIE extractions may impact the overall quality of our data generation framework.} As illustrated in the top half of Figure~\ref{fig:falsesum_input}, we extract the relation tuples from each source document and its corresponding reference summaries.
To minimize the risk of $\mathcal{G}$ inadvertently generating consistent summaries, we corrupt each extracted ``fact'' by removing one randomly chosen argument from each tuple. For instance, OpenIE may extract the following tuple from a sentence:
$$
(\frac{\texttt{Jo}}{\tiny\texttt{ARG}_0}, \frac{\texttt{\textbf{plans to give}}}{\tiny\texttt{PRED}}, 
\frac{\texttt{Alex}}{\tiny\texttt{ARG}_1},
\frac{\texttt{apples}}{\tiny\texttt{ARG}_2})
$$
We then randomly choose $\texttt{apples}_{\tiny\textsc{ARG}_2}$ to be removed from the tuple. We additionally lemmatize the dependency root word of each argument and predicate span, e.g., $\textbf{plans to give}\Rightarrow \textbf{plan to give}$. 
This forces the model to learn to correct for grammaticality by inflecting the spans when inserting them to the \textbf{masked} spans. Once all such spans are extracted and processed, they are \textbf{grouped} and \textbf{shuffled} into two lists (predicates and arguments).

\begin{table*}
\centering
\footnotesize
\begin{tabular}{p{1.1cm}|p{7.8cm}|p{3.2cm}|p{2.2cm}}
\toprule
     \textbf{Mode} & \textbf{Input} & \textbf{Expected Output} & \textbf{Description} \\
\midrule
\textbf{train} \quad intrinsic & 
\predbox{$\texttt{Predicates}$}: \hlpred{caught}, \hlpred{\textbf{plead guilty to}}, \hlpred{$\dots$}, \hlpred{appear before}, \hlpred{face}; \argbox{$\texttt{Arguments}$}: \hlarg{the corruption scandal}, \hlarg{\textbf{Two Pennsylvania judges}}, \hlarg{$\dots$}, \hlarg{many children}, \hlarg{the U.S.} \codebox{$\texttt{Code}$}: \hlcode{intrinsic}; \sumbox{$\texttt{Summary}$}:\hlblank{1} \hlblank{0} \hlsum{federal fraud charges.}
& \hlarg{Two Pennsylvania judges} \hlarg{plead guilty to} federal fraud charges. 
& Model learns to combines listed spans to produce most plausible summary. \\
\midrule
\textbf{test} \quad intrinsic & 
\predbox{$\texttt{Predicates}$}: \hlpred{caught}, \semitransp{plead guilty to}, \hlpred{$\dots$}, \hlpred{appear before}, \hlpred{face}; \argbox{$\texttt{Arguments}$}: \hlarg{the corruption scandal}, \semitransp{Two Pennsylvania judges}, \hlarg{$\dots$}, \hlarg{many children}, \hlarg{the U.S.} \codebox{$\texttt{Code}$}: \hlcode{intrinsic}; \sumbox{$\texttt{Summary}$}:\hlblank{1} \hlblank{0} \hlsum{federal fraud charges.}
& \hlerror{Many of the children} \hlerror{face}  federal fraud charges.
& Model consolidates incorrect information. \\
\midrule
\midrule
\textbf{train} \quad extrinsic & 
\predbox{$\texttt{Predicates}$}: \semitransp{is pressing for}, \hlpred{limit}, \hlpred{$\dots$}, \hlpred{is being erode}, \hlpred{is fight}; \argbox{$\texttt{Arguments}$}: \hlarg{panelist}, \semitransp{action}, \hlarg{$\dots$}, \hlarg{sea level}, \hlarg{Arctic melt}, \semitransp{at the climate change conference} \codebox{$\texttt{Code}$}: \hlcode{extrinsic}; \sumbox{$\texttt{Summary}$}: \hlsum{The Alliance} \hlblank{0} \hlblank{1} \hlblank{2}.
& The Alliance \hlarg{is pressing for} \hlarg{action} \hlarg{at the climate change conference}.
& Model learns to hallucinate new unsupported information. \\
\midrule
\textbf{test} \quad extrinsic & 
\predbox{$\texttt{Predicates}$}: \semitransp{is pressing for}, \hlpred{limit}, \hlpred{$\dots$}, \hlpred{is being erode}, \hlpred{is fight}; \argbox{$\texttt{Arguments}$}: \hlarg{panelist}, \semitransp{action}, \hlarg{$\dots$}, \hlarg{sea level}, \hlarg{Arctic melt}, \semitransp{at the climate change conference} \codebox{$\texttt{Code}$}: \hlcode{extrinsic}; \sumbox{$\texttt{Summary}$}: \hlsum{The Alliance} \hlblank{0} \hlblank{1} \hlblank{2}.
& The Alliance \hlerror{is planning to} \hlerror{impose} \hlerror{limits on emissions}.
& Model hallucinates new unsupported information. \\
\bottomrule          
\end{tabular}
\caption{Examples of input formatting on two different summarization instances for both intrinsic and extrinsic error types during training and testing. Gold input spans (indicated by \textbf{boldface}), which are extracted from the gold summary, are only visible to the model during intrinsic training. They are removed from the input in all other settings, as indicated by \st{strikethrough} text.}
\label{tab:input_format}
\end{table*}

\subsection{Input Formatting} 
\label{sec:input-formatting}
Let $\mathrm{P} =$ $(\texttt{PRED}_1, \dots, \texttt{PRED}_n)$ and $\mathrm{A} =$ $(\texttt{ARG}_1,$ $\dots,$ $\texttt{ARG}_m)$ be the unordered lists of extracted predicates and arguments from a source document $\mathrm{D}$ and the summary sentence $\mathrm{S}^+$. Additionally, we assume a \textbf{masked} summary sentence $\mathrm{M}$ (described later), derived from $\mathrm{S}^+$, and a control code variable $c \in \{\texttt{intrinsic}, \texttt{extrinsic}\}$. Generator $\mathcal{G}$ is trained to compute $p(\mathrm{S}^+ | \mathrm{P}, \mathrm{A}, \mathrm{M}, c)$. As illustrated in the bottom half of Figure~\ref{fig:falsesum_input}, we encode all the conditional variables into the following format:
$$
\footnotesize
\texttt{Predicates:} \mathrm{P};\texttt{Arguments:} \mathrm{A}; \texttt{Code:} c; \texttt{Summary:} \mathrm{M}
$$
In the following, we describe the key steps in the input formatting process:

\paragraph{Step 1: Span Removal} Initially, $\mathrm{P}$ and $\mathrm{A}$ include predicate and argument spans from the original summary which may be used to reconstruct $\mathrm{S}^+$. However, at \textbf{test} time we remove these ``gold'' spans from the two lists to force the $\mathcal{G}$ to make consistency mistakes. The removal is also done when training the model for control code $\texttt{extrinsic}$ to train $\mathcal{G}$ to predict plausible unseen spans.\footnote{It is possible that some spans from the source document are duplicates of gold ones. For instance, the document may mention ``The Queen of England'', while the gold span from the summary is ``The Queen''. We use a simple heuristic to remove such duplicates by searching for other spans whose (lemmatized) dependency root token is the same.} We summarize the different input formatting in Table~\ref{tab:input_format}. 

\paragraph{Step 2: Span Reduction} To encourage $\mathcal{G}$ to generate fine-grained errors~\cite{pagnoni-etal-2021-understanding, goyal-durrett-2021-annotating}, we also train it to hallucinate incorrect modifiers into spans from $\mathrm{P}$ and $\mathrm{A}$. To this end, we randomly drop adjectives and adverbs from $10\%$ of the gold predicate and argument spans.
For instance, an argument span ``recently elected prime minister'' will be reduced to ``minister''. This teaches the model to generate the remaining part of the span given only the context provided in the formatted input.

\paragraph{Step 3: Control Code} To control the type of consistency errors generated by $\mathcal{G}$, we append the string ``$\texttt{code:}$'' followed by either ``$\texttt{intrinsic}$'' or ``$\texttt{extrinsic}$'' into the input tokens. The code is determined randomly with equal probability of $0.5$. Once the code is chosen, we perform the remaining formatting steps accordingly (see Table~\ref{tab:input_format}).

\paragraph{Step 4: Summary Masking} We derive masked summary $\mathrm{M}$ by replacing the spans of \textbf{randomly} selected predicates and arguments with a special token $\texttt{\textbf{<span\char`_i>}}$, where $i=0$ is reserved for the predicate, and $i>0$ for their arguments. These tokens control \textbf{where} the incorrect information should be inserted by the generator model into the original summary (see Table~\ref{tab:input_format}).

\subsection{Training \falsesum{}} We run the \falsesum{} data generation pipeline on the \textit{train} split of the CNN/DailyMail corpus~\cite{hermann-etal-2015-teaching}, originally collected for question answering, but subsequently reformulated for summarization by~\citet{nallapati-etal-2016-abstractive-2}. This dataset contains \emph{English} news documents paired with human-written summaries, each consisting of multiple sentences. We break the summaries down such that each \falsesum{} example consists of the document text and a single sentence summary. We then run the \textbf{preprocessing} and \textbf{formatting} steps on each document-summary pair. The resulting pairs of formatted input and target output are subsequently split into train and test sets which consist of 394,774 and 262,692 instances, respectively. 

We use the $\texttt{T5-base}$ model~\cite{raffel-etal-2020-exploring} as generator $\mathcal{G}$ and fine-tune it on the seq2seq task described in \sect{sec:design-overview}. The NLI examples are produced by running the fine-tuned generator on the preprocessed and formatted test split.\footnote{See Appendix~\ref{sec:app_hyperparam} for the hyperparameter details.} This renders an equal number of positive and negative examples. In our experiments, we randomly sample 100,000 \falsesum{} examples to augment the NLI dataset.

\section{Experimental Settings}
Our experiments aim to demonstrate the effectiveness of \falsesum{}-generated document-level examples for NLI dataset augmentation. We evaluate the downstream performance of the NLI models by testing them against several benchmarks for determining the factual inconsistency of generated summaries. In this section, we describe the training setup of the NLI models, including the model and both the sentence- and document-level datasets.

\subsection{Training}

\paragraph{NLI models} We train several NLI models by fine-tuning $\texttt{RoBERTa-base}$~\cite{liu-etal-2018-roberta} on \emph{either} the original or the augmented MNLI dataset~\cite{williams-etal-2018-broad}. The MNLI dataset consists of 392,702 train instances, each labeled as either ``\textit{entailment}'', ``\textit{neutral}'', or ``\textit{contradiction}''. To enable the application of NLI data to this factual consistency task, we use a binary formulation of NLI, where the \textit{``neutral''} and \textit{``contradiction''} labels are combined into \textit{``non-entailment''}. The document-level inputs are formatted similarly to sentence-level examples, i.e., the document premise $\mathrm{D}$ and summary hypothesis ($\mathrm{S}^+$ or $\mathrm{S}^-$) are concatenated and a special classification token ($\texttt{[CLS]}$) is used~\cite{devlin-etal-2019-bert}.

\paragraph{Document-level NLI datasets} We conduct augmentation comparisons with several multi-sentence NLI datasets which obtain examples from \emph{news} or \emph{summarization} domains. We consider the following datasets: \textbf{ANLI}~\cite{nie-etal-2020-improving}, a paragraph-level NLI dataset collected via an iterative and adversarial human-in-the-loop annotation protocol. It consists of mostly Wiki data but also includes a small portion of news text; \textbf{DocNLI}~\cite{yin-etal-2021-docnli}, a document-level NLI dataset containing multi-sentence premise and hypothesis sentences, collected by converting QA examples to NLI instances~\cite{demszky-etal-2018-transforming} and replacing words and sentences in \emph{news} summaries using a language model; \textbf{\factcc{}}~\cite{kryscinski-etal-2020-evaluating}, a large-scale dataset specifically generated for training summary factual correctness classification models. The positive examples in \factcc{} are obtained by backtranslating a random sentence from a CNN/DailyMail \emph{news} story, while negative examples are obtained by perturbing the sentence using predefined rules, e.g., entity swapping. For fair comparison, we sample 100,000 examples from each augmentation dataset in our experiments.

\begin{table*}
\centering
\small
\begin{tabular}{r|l|cccc|c}
\toprule
 & & \multicolumn{4}{c|}{\textbf{Benchmark Datasets}} \\
 \textbf{Dataset} & \textbf{Augmentation} & \textbf{\factcc{}} & \textbf{Ranksum} & \textbf{QAGS} & \textbf{SummEval} & \textbf{Overall}  \\
\midrule
\textit{Majority voting} &  - & 50.00 & 50.46 & 50.00 & 50.00 & 50.11  \\
\midrule
\textbf{MNLI}-128 & - & 57.39 & 57.01 & 59.72 & 54.11 & 57.06  \\
$\texttt{[split-doc]}$ \textbf{MNLI}-128 & - & 72.07 & 68.03 & 71.08 & 55.32 & 66.63 \\
\midrule
\textbf{MNLI}-512 & - & 57.93 & 51.40 & 52.73 & 48.75 & 51.43  \\
\textbf{MNLI}-512 & $\texttt{ANLI}$ & 53.91 & 55.76 & 53.54 & 49.56 & 53.19  \\
\textbf{MNLI}-512 & $\texttt{DocNLI}$ & 58.13 & 53.58 & 57.10 & 52.59 & 55.35  \\
\textbf{MNLI}-512 & $\texttt{\factcc{}}$ & 73.87 & 67.29 & 73.50 & 60.04 & 69.02  \\
\textbf{MNLI}-512 & \falsesum{} (ours) & \textbf{83.52} & \textbf{72.90} & \textbf{75.05} & \textbf{65.18} & \textbf{74.17}  \\
\bottomrule          
\end{tabular}
\caption{Performance of MNLI models with different augmentation data across benchmarks to classify the factual consistency of summaries. \textbf{MNLI}-128 and \textbf{MNLI}-512 are $\texttt{RoBERTa-base}$ models trained using maximum token length of 128 and 512, respectively.}
\label{tab:results_all}
\end{table*}

\subsection{Benchmark Datasets}
We evaluate these NLI models on four benchmark datasets to classify the factual consistency of abstractive summaries. These datasets differ in terms of the annotation protocol, the granularity of the summaries (single- or multi-sentence), the summarization corpus used, and the models used to generate the summaries that are annotated. The tasks are formulated as a binary classification with the labels \textit{``consistent''} and \textit{``inconsistent''}. We evaluate NLI models on these tasks by mapping the predicted label \textit{``entailment''} to \textit{``consistent''} and \textit{``non-entailment''} to \textit{``inconsistent''}.
The benchmarks datasets are detailed in the following:

\paragraph{\factcc{}} In addition introducing a synthetic training dataset for the task,~\citet{kryscinski-etal-2020-evaluating} introduce a manually annotated test set. It contains 1,431 document and single-sentence summary pairs generated by various neural abstractive summarization models trained on CNN/DailyMail corpus.\footnote{We merge the test and validation sets into a single test set.} 

\paragraph{Ranksum}~\citet{falke-etal-2019-ranking} formulate the factual consistency problem in summarization as a ranking task. They introduce a dataset consisting of 107 documents, each paired with a set of five ranked summary candidates obtained from the beam search of a summarization model. Given the manually annotated consistency label on summary candidates, the task is to re-rank the list such that the top-1 summary is factually consistent. 

\paragraph{Summeval} ~\citet{fabbri-etal-2021-summeval} introduce a comprehensive benchmark for factual consistency detection in summarization. It includes summaries generated by seven extractive models and sixteen abstractive models, which are judged by three annotators using a 5-point Likert scale.\footnote{We aggregate the label as ``consistent'' if all annotators rated the summary as a 5 and ``inconsistent'' otherwise.}

\paragraph{QAGS} The dataset collected by~\citet{wang-etal-2020-asking} consists of 239 test set instances from XSUM~\cite{narayan-etal-2018-dont} and 714 instances from CNN/DailyMail.\footnote{This is the number of instances after we split multi-sentence summaries into separate single-sentence summary test instances, where an individual factuality judgement is available.} Each instance consists of a pair of a source document and a single-sentence summary, which is labeled via majority voting on three annotators' labels.

\section{Results and Discussion}
\subsection{Main Results}
Performance on \factcc{}, QAGS, and SummEval is measured using balanced accuracy, which is suitable for class imbalanced settings, since the factually consistent label is the majority in some benchmark datasets. It is defined as the average recall of the two classes, such that majority label voting obtains only a 50\% score. To measure ranking performance in Ranksum, we calculate the average Precision@1, which computes the fraction of times a factually consistent summary is ranked highest on each test instance. We perform five training runs for each setup using different random seeds and take the mean to address performance instability~\cite{reimers-gurevych-2017-reporting}.

From the results in Table~\ref{tab:results_all}, we observe the following: 
\textbf{(1)} Models trained on sentence-level MNLI datasets perform poorly when evaluated directly on document-level benchmarks, even after we increase the maximum input token length from 128 to 512;\footnote{Average context word count is only 22 in MNLI and 546 in \factcc{}.}
\textbf{(2)} This limitation can be alleviated by the sentence-wise prediction strategy ($\texttt{[split-doc]}$MNLI-128),\footnote{See details in Appendix~\ref{sec:app_aggregating}} which achieves 66.63. Note, however, that this improvement comes at the expense of compute cost which is multiplied by a significant factor; \textbf{(3)} DocNLI and ANLI perform poorly even though they contain longer premise sentences, indicating that the length mismatch may not be the primary issue; \textbf{(4)} \falsesum{} obtains substantial improvement over the previous state-of-the-art \factcc{}, despite being derived from the same summarization dataset (CNN/DailyMail). This indicates that \falsesum{} provides higher quality examples and includes more types of entailment phenomena that occur naturally in this task.

\begin{table}
\centering
\small
\begin{tabular}{l|cc}
\toprule
\textbf{Training Dataset} & \textbf{Overall} & \textbf{$\Delta$} \\
\midrule
MNLI+\falsesum{} &  \textbf{74.17} &  \\
MNLI+\falsesum{} $\texttt{-Contrastive}$ &  73.11 & {\color{purple} -1.06} \\
MNLI+\falsesum{} $\texttt{-Extrinsic}$ &  71.95 & {\color{purple} -2.22} \\
MNLI+\falsesum{} $\texttt{-Intrinsic}$ &  69.14 & {\color{purple} \textbf{-5.03}} \\
\bottomrule          
\end{tabular}
\caption{Model performance when trained on ablated \falsesum{} dataset. Excluding the contrastive, extrinsic, and intrinsic examples results in lower overall performance, indicating each property is beneficial.}
\label{tab:results_ablation}
\end{table}

\subsection{Ablation Analysis on \falsesum{} Data}
We perform an ablation analysis to study how each component of our data generation pipeline contributes to the final performance. We first remove the contrastive property of the \falsesum{} data by randomly including only \textbf{either} the positive $(\mathrm{D}, \mathrm{S}^+, Y=1)$ \textbf{or} negative $(\mathrm{D}, \mathrm{S}^-, Y=0)$ NLI examples obtained from a single  $(\mathrm{D}, \mathrm{S}^+)$ pair. Next, we filter out the negative NLI instances that are generated using either $\texttt{intrinsic}$ or $\texttt{extrinsic}$ code. We refer to the three ablated datasets as $-\texttt{contrastive}$, $-\texttt{intrinsic}$ and $-\texttt{extrinsic}$, respectively. We set the sampled training size to 100,000 for the three ablation setups and aggregate the results from five training runs.

Table~\ref{tab:results_ablation} shows the performance of the ablated models. We observe that removing contrastive pairs in the augmented training data results in a $1.06\%$ drop on the overall benchmarks score. We also see that removing $\texttt{intrinsic}$ error examples results in the highest performance loss, $-5.03\%$ compared to $-2.22\%$  by $-\texttt{extrinsic}$. This is explained by the fact that intrinsic consistency errors are more dominant on benchmarks that are built on the CNN/DailyMail corpus~\cite{goyal-durrett-2021-annotating}. We conclude that all the above properties are important for the overall improvements obtained by \falsesum{}.

\begin{figure}
    \centering
    \includegraphics[width=0.95\linewidth]{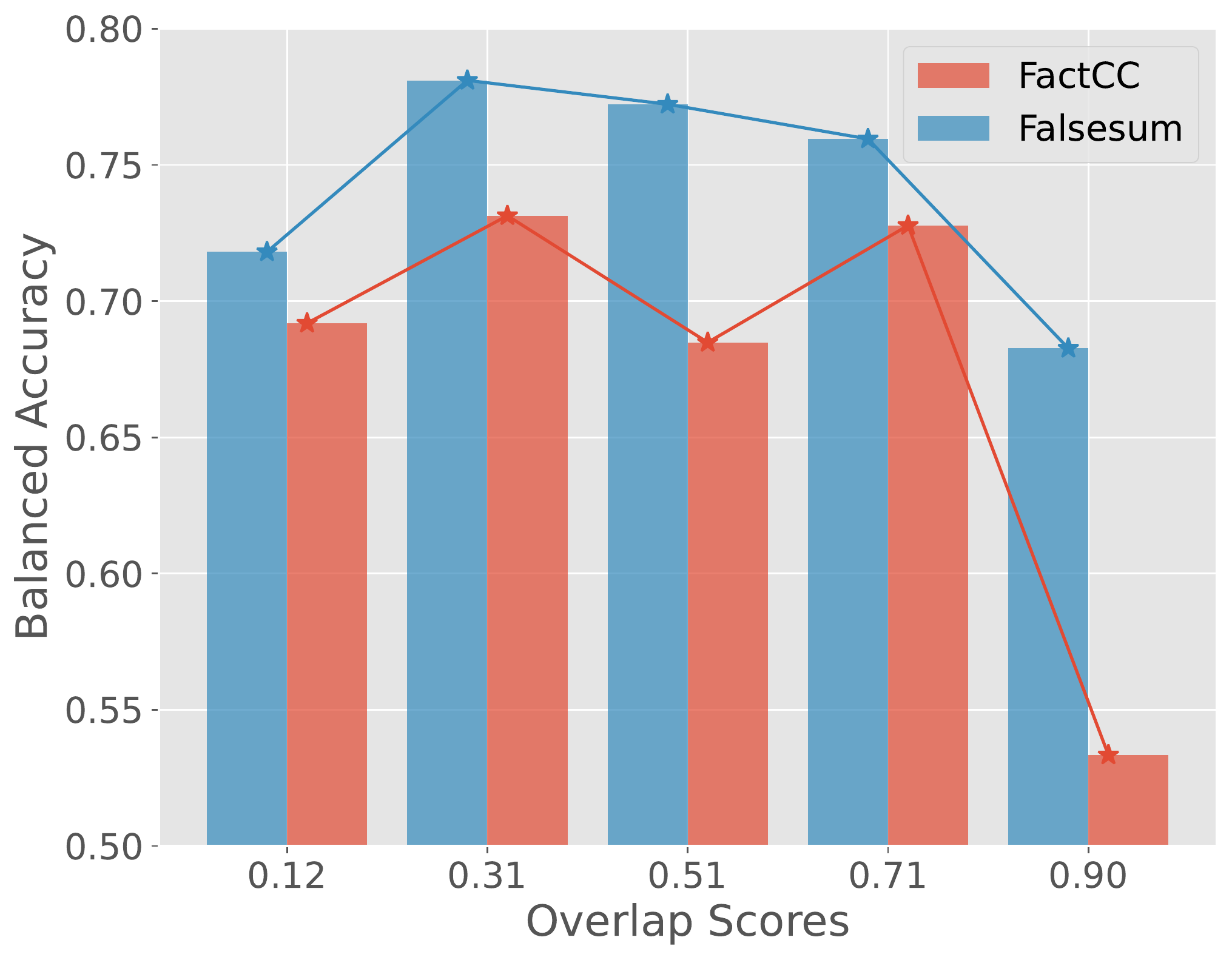}
    \caption{Comparison between NLI models augmented with \falsesum{} and \factcc{} across different measures of summary extractiveness. The x-axis shows the median overlap score of each test subset.}
\label{fig:extractiveness_evaluation}
\end{figure}

\subsection{Fine-grained Evaluation}
Previous work has shown that NLI models are prone to relying on fallible heuristics which associate lexical overlap with entailment labels~\cite{mccoy-etal-2019-right}. In the factual consistency task, this corresponds to models associating highly extractive summaries with the ``consistent'' label. This raises a question about whether \falsesum{} data alleviates this tendency in the resulting NLI models.

To answer this question, we partition the \factcc{} annotated test examples into five ordered subsets based on the lexical overlap between their summary hypothesis and the source document premise. We define an overlap score using the $\textsc{normalized coverage}$ and $\textsc{density}$ summary extractiveness scores introduced by~\citet{grusky-etal-2018-newsroom}. Both measures have the range $[0.0, 1.0]$, where $\textsc{density}=1.0$ indicates that all words in a summary are also present in the source document and $\textsc{normalized coverage}=1.0$ indicates that the summary is obtained by copying a continuous fragment of the source document. We then define $\textsc{overlap}=\textsc{normalized coverage} \times \textsc{density}$.

Figure~\ref{fig:extractiveness_evaluation} shows the comparison of \factcc{} and \falsesum{} augmentation performance across varying lexical overlap scores. We see that \falsesum{} performs better on all subsets of the \factcc{} test set with the greatest performance gap appearing on the $0.9$ overlap subset. Upon closer inspection, we see that the \factcc{} model makes mostly false positive classification errors on this subset, i.e., it tends to predict highly extractive summaries as ``consistent'', leading to near majority voting performance of $50\%$. \falsesum{}, on the other hand, better discriminates the factual consistency of examples without over-relying on lexical overlap.

\subsection{Data Quality Analysis}
We conduct both manual and automatic quality evaluation of the \falsesum{}-generated dataset. First, we sample 200 generated negative examples and manually verify whether (i) the perturbed summary $\mathrm{S}^-$ is indeed factually inconsistent; (ii) the type of consistency error follows the specified control code; (iii) the incorrect ``fact'' is inserted at the specified missing span. Following~\citet{kryscinski-etal-2020-evaluating}, the authors perform this annotation to avoid high disagreement by crowd annotators in this task~\cite{falke-etal-2019-ranking}. The results in Table~\ref{tab:manual_verification} show that about $86\%$ of intrinsic $81\%$ of extrinsic generated error examples are factually inconsistent, which happen due to several reasons, e.g., generator model chooses a span from the list that is similar to the original span, or generator model correctly guesses the original missing span.
This further suggests that pre-trained language models such as $\texttt{RoBERTa-base}$ can be robust against the induced label noise and can still learn a performant classifier. While $\mathcal{G}$ almost always inserts the incorrect ``fact'' at the specified positions, we observe that it often fails to follow the specified extrinsic code correctly. We suspect that this is because the model prefers the easier task of copying the input over generating novel phrases.\footnote{We include more examples of generated NLI instances as well as the inadvertently consistent output in Appendix~\ref{sec:app_more_examples}.}

\begin{table}
\centering
\footnotesize
\begin{tabular}{l|l|l|l}
\toprule
\textbf{Code} & \textbf{Label} \checkmark & \textbf{Type} \checkmark & \textbf{Span} \checkmark \\
\midrule
Intrinsic & 86\% & 94\% & 94\% \\
Extrinsic & 81\% & 65\% & 95\% \\
\bottomrule          
\end{tabular}
\caption{Manual verification of \falsesum{}-generated NLI examples. Label, type, and span indicate the percentage of generated summaries with correct inconsistency label, error type, and error span, respectively.}
\label{tab:manual_verification}
\end{table}

Following~\citet{gururangan-etal-2018-annotation}, we also evaluate the naturalness of the generated dataset. We train an NLI model using positive examples from CNN/DailyMail and \falsesum{}-generated negative examples. The model receives no premise so must distinguish between entailed and non-entailed hypotheses using semantic plausibility or spurious surface features, e.g., grammatical mistakes or fluency errors. The relatively low accuracy of these models on \falsesum{} data (shown in Table~\ref{tab:artifacts_analysis}) suggests that, compared to \factcc{} and DocNLI, \falsesum{}-generated summaries are relatively hard to distinguish from the gold ones.

\begin{table}
\centering
\footnotesize
\begin{tabular}{l|l|l|l}
\toprule
& \textbf{\factcc{}} & \textbf{DocNLI} & \textbf{\falsesum{}}  \\
\midrule
Majority voting & 50.84 & 53.55 & 50.00 \\
\midrule
CBOW-GloVe & 60.36 & 70.38 & 56.13 \\
BiLSTM-GloVe & 68.26 & 73.04 & 57.62 \\
RoBERTA-base & 82.15 & 78.46 & 69.38\\
\bottomrule          
\end{tabular}
\caption{Hypothesis-only model performance (accuracy) to measure the presence of artifacts and naturalness of \falsesum{} dataset (lower is better).}
\label{tab:artifacts_analysis}
\end{table}

\section*{Conclusion}
NLI models present a promising solution for automatic assessment of factual consistency in summarization. However, the application of existing models for this task is hindered by several challenges, such as the mismatch of characteristics between their training dataset and the target task data. This mismatch includes the difference in terms of the input granularity (sentence vs. document level premises) and the types of (non-)entailment phenomena that must be recognized.

In this work, we present \falsesum{}, a data generation pipeline which renders large-scale document-level NLI datasets without manual annotation. Using our training strategy, we demonstrate that it is possible to learn to generate diverse and naturalistic factually inconsistent (non-entailed) summaries using only existing (entailed) consistent summaries for training. We show that the resultant data is effective for augmenting NLI datasets to improve the state-of-the-art performance across four summary factual inconsistency benchmarks. 

\section*{Acknowledgments}
We would like to thank Marco Ponza, Marco Fiscato, Umut Topkara and other colleagues from Bloomberg AI for the thoughtful discussion and feedback throughout this project. We also thank Leonardo Ribeiro for comments on the earlier version of this work and the anonymous reviewers for their constructive feedback. The authors affiliated with UKP were supported by the German Research Foundation through the research training group “Adaptive Preparation of Information from Heterogeneous Sources” (AIPHES, GRK 1994/1) and by the German Federal Ministry of Education and Research and the Hessian State Ministry for Higher Education, Research and the Arts within their joint support of the National Research Center for Applied Cybersecurity ATHENE.

\bibliography{anthology,custom}
\bibliographystyle{acl_natbib}

\clearpage
\appendix

\section{Hyperparameters}
\label{sec:app_hyperparam}
\paragraph{Generator model} We train a $\texttt{T5-base}$ model for three epochs with batch size of 24 using the AdamW optimizer. We set the maximum source token length to 256 and the target token length to 42. We use a learning rate of $3e^{-5}$ and fix the random seed to $11$. For decoding, we set the minimum and maximum sequence length to 10 and 60, respectively. We sample using beam search with a beam of size two. We additionally set the repetition penalty to 2.5 and the length penalty to 1.0.
\paragraph{Classification model} We train $\texttt{RoBERTa-base}$ models on augmented and original MNLI datasets for three epochs with a batch size of 32. The learning rate is set to $1e^{-5}$, while the maximum input token length is set to either 128 or 512. We use the following random seeds for the five training runs: 11, 12, 13, 14, and 15.

\section{Aggregating Predictions}
\label{sec:app_aggregating}
We follow~\citet{falke-etal-2019-ranking} to adapt out-of-the-box MNLI models to document-level input by performing a sentence-wise prediction before aggregating the output. Given a document $\mathrm{D}$ consisting of sentences $d_1, \dots, d_n$, and a multi-sentence summary $S$ consisting of $s_1, \dots, s_m$, we aggregate the probability scores given by the classifier model $F$ on each $d_i, s_j$ pair. The aggregated consistency score $\sigma(\mathrm{D}, S)$ is given by:
$$
\sigma(\mathrm{D}, S) = \frac{1}{m} \sum_{j=1}^{m} \argmax_{d \in \mathrm{D}} F(d, s_j)
$$
This means that it is sufficient for a summary sentence to be factually consistent given only a single entailing sentence in the source document. We then take the average scores across the summary sentences since each of them needs to be entailed by the source document. We use a similar aggregation method to evaluate augmented MNLI models on multi-sentence summaries from the Summeval and Ranksum benchmarks.

\section{\falsesum{} Details}
In the preprocessing steps, we only perform the predicate and argument span extraction on the first 15 sentences for computational efficiency. For training, this is not an issue since the gold spans from the reference summary are included in the input. Additionally, we may extract multiple OpenIE relation tuples from each sentence. To avoid having overlapping spans from a single input, we randomly select two tuples from each sentence.

\section{Falsesum Examples}
\label{sec:app_more_examples}
We include more examples of generated NLI instances in Table~\ref{tab:falsesum_nli_examples}. We also include cases where \falsesum{} inadvertently generates factually consistent summaries in Table~\ref{tab:fail_cases}. Lastly, we show several examples of the formatted input and the generated output at \textbf{test} time in Table~\ref{tab:input_output_examples}.

\include{examples}

\end{document}

%% file: examples.tex
\begin{table*}
\centering
\footnotesize
\begin{tabular}{rl}
\toprule
\multicolumn{2}{p{16cm}}{
    Mexican federal police have arrested a fugitive on the FBI's 10 Most Wanted list, Mexican authorities said. Jorge Alberto Lopez Orozco allegedly murdered his girlfriend and her two young sons. Jorge Alberto Lopez Orozco is wanted in Elmore County, Idaho, on charges that he shot and killed three people, the FBI said. The charred remains of a woman and her sons, ages 2 and 4, were found inside a burned-out vehicle on August 11, 2002, it said. Each victim had been shot in the head or chest. The \hlarg{FBI was still working Friday to confirm the identity of the man in custody}, said Debbie Dujanovic, a spokeswoman in the agency's Salt Lake City, Utah, field office. The Salt Lake City office has jurisdiction in the case. \hlerror{An extradition order was issued in} January 2007, the Mexican attorney general's office said in a news release Thursday. A reward of up to \$100,000 was being offered, the FBI said. Lopez, 33, was captured in Zihuatanejo, a city northwest of Acapulco on the Pacific Coast in southern Mexico, the Mexican attorney general's office said. Zihuatanejo is in Guerrero state, but Lopez was transferred to a jail in neighboring Michoacan state, officials said. The arrest came about after investigation and intelligence work by Mexican authorities, the attorney general's office said. According to the FBI, Lopez abducted his girlfriend, Rebecca Ramirez, and her two young sons from her father's house in Nyssa, Oregon, on \hlerror{July 30, 2002}. The car he had been driving was found nearly two weeks later on a rural road near Mountain Home, Idaho, officials said. $\dots$
}\\
\midrule
    entailment & FBI was still working Friday to confirm the identity of the man in custody. \\
    (intrinsic) non-entailment & \hlerror{An extradition order was issued in July 30, 2002, to determine} the identity of the man in custody. \\
\midrule
\multicolumn{2}{p{16cm}}{
    He may have been allowed to leave the club without ever playing a league game for the first team, but \hlarg{Kristoffer Olsson still showed Arsenal some love as he departed.} The 19-year-old Swede, whose only first-team appearance for the Gunners came off the bench in the Capital One Cup last season, has joined FC Midtjylland this week on a permanent deal. But, as the news was announced, \hlarg{Olsson took to Twitter to say 'Once a Gunner, always a Gunner'}. Kristoffer Olsson (right) played just once for Arsenal's first team, in the Capital One cup against West Brom . Olsson expressed his love for the club on Twitter, despite being sold to FC Midtjylland . The tweet reflects Cesc Fabregas' comments when he left the club to join Barcelona, although the Spanish midfielder has sinced joined rivals \hlerror{Chelsea}, after Arsene Wenger opted not to buy him back. \hlerror{Olsson has been on loan at FC Midtjylland} since the beginning of the season, playing six times in the Danish top flight. The Sweden U21 international said on joining permanently: 'this is a club that believes in me and sees my potential.' Olsson has played six times on loan with FC Midtjylland and has now joined the Danish club permanently.
}\\
\midrule
    entailment & Swedish international takes to social media to express love for Arsenal. \\
    (intrinsic) non-entailment & Swedish international \hlerror{has been on loan at Chelsea since last season}. \\
\midrule
\multicolumn{2}{p{16cm}}{
    A teenager who was struck down with an agonising bowel condition says dancing has helped him to overcome his debilitating illness. \hlarg{Macaulay Selwood, 17, was diagnosed with Crohn's} two years ago and was so unwell that he was often left in agony on the floor unable to move. But his determination to continue his promising dancing career gave him the spur he needed to battle through. Lord of the Dance: Macaulay at his practice studio. \hlarg{He was diagnosed with Crohn's in September 2010} after collapsing in agony during a dance class . Recovery: 'Dancing has helped me overcome it (Crohn's). It kept me motivated' Now the teenager from Bristol has made it to the finals of the Irish dancing world championships in Boston, USA, and is hotly-tipped for glory. He will then have a trial at the famous performing arts school, ArtsEd, in London. At shows he has been compared with Riverdance star Michael Flatley while others have taken to calling him Billy Elliot, after the film character who overcomes the odd to becoming a dancing star. Macaulay did ballet at college before focusing on Irish dancing for the world championships and works at Tesco to fund his passion. $\dots$
}\\
\midrule
    entailment & Macaulay Selwood, 17, first starting suffering from Crohn's disease in 2010. \\
    (extrinsic) non-entailment & \hlerror{The 22-year-old}, who was diagnosed with Crohn's in 2010, \hlerror{has been recovering since} 2010. \\
\midrule
\multicolumn{2}{p{16cm}}{
    When Matthew Briggs, 32, from Huntington in North Yorkshire noticed that his father had posted a photo of them together on Facebook, he was initially pleased. But when he opened the photo and saw the image, Mr Briggs was left horrified by the sight of his 31st frame. Now, two years on, he has shed an astonishing 17st and, in November, will complete the New York marathon in memory of his mother Susan who died from multiple sclerosis when he was just 18. Pounding the pavements: Matthew Briggs, 32, has lost an impressive 17st in just two years of slimming . \hlarg{'In March of 2000, she lost her battle with Multiple Sclerosis,'} he says. 'She has always been my inspiration. I am the man I am today because of the woman she was.' \hlarg{Money raised by Mr Briggs' 26-mile run will be donated to the Multiple Sclerosis Society, a charity dedicated to beating the disease} as well as supporting sufferers and their families. Mr Briggs, who has dropped from 31st to just under 14st, had piled on the pounds thanks to a diet of ready meals, takeaways and daily two litre bottles of Coca-Cola. But, after seeing the photo posted on Facebook and spurred on by a bet with his father, Mr Briggs joined his local Slimming World group and went on to shed more than 17st over two years. $\dots$
}\\
\midrule
    entailment & She died in 2000 of multiple sclerosis and funds raised will go to charity. \\
    (extrinsic) non-entailment & She died in 2000 of multiple sclerosis and \hlerror{every penny she saves} will go to charity. \\
\bottomrule          
\end{tabular}
\caption{Examples of NLI pairs generated by \falsesum{}. We show both the entailment and non-entailment hypotheses obtained from each source document. \hlarg{Green-highlighted} spans indicate the information used consistently in the summary. \hlerror{Red-highlighted} spans indicate information used or inserted by the model to generate an inconsistent summary.}
\label{tab:falsesum_nli_examples}
\end{table*}

\begin{table*}
\centering
\footnotesize
\begin{tabular}{rl}
\toprule
\multicolumn{2}{p{16cm}}{
    The Mojito, a Cuban mix of white rum, sugar, lime, mint and soda water, is the most popular cocktail in Britain according to a report . \hlarg{Sales of cocktails have risen by more than 10 per cent in the past two years}. More than one in five of Britain's pubs and bars now serve cocktails and the Mojito – a Cuban mix of white rum, sugar, lime, mint and soda water – is the most popular, according to a report. Pina Coladas (rum, coconut and pineapple juice) and Woo Woos (vodka, peach schnapps and cranberry juice) were also popular. The Mixed Drinks Report, by consultancy firm CGA Strategy, found more women than men choose cocktails, as 54 per cent of cocktail drinkers are female. Bomb and pitcher serves remain popular, with 74 per cent of 18 to 24-year-olds admitting to have bought a bomb drink, while nine in 10 in the same age range say they drink pitchers. Cocktails are enjoyed by the core 18 to 35-year-old demographic 'in all on-trade occasions' including throughout the night, as opposed to just the start. $\dots$
}\\
\midrule
    gold & Sales of cocktails have risen by more than 10 per cent in the past two years. \\
    (extrinsic) generated & Cocktails \hlarg{have soared in popularity over} the past two years. \\
\midrule
\multicolumn{2}{p{16cm}}{
    From Yellowstone National Park to the Everglades, \hlarg{America's 391 national parks are in need of repair} -- and thanks to the economic stimulus signed into law, help is now underway. President Obama and his family visit the Grand Canyon in Arizona, a national park. President Obama's \$787 billion economic stimulus plan passed in February and designated \$750 million dollars to the national parks. But not all of the stimulus money is being used -- \hlarg{and the parks are facing a \$9 billion backlog in maintenance projects}. So far, nearly 10 percent is in the pipeline. "We are picking away at it as much as we can and we've been fortunate to have the recovery act money," said Jeffrey Olson of the National Park Service. Olson said half of the \$9 billion is slated to go for road repairs. "Half of that [\$9 billion] is roads and about \$2 billion of that are the most pressing needs -- those we get some help from the stimulus. The president's budget proposal is calling for more maintenance and construction money," Olsen said. Dan Wenk, the acting director of the National Park Service says most of those pressing needs include, "camp grounds, camp sites, it's amphitheaters for evening programs. It's the bathrooms. $\dots$
}\\
\midrule
    gold & Park Service is dealing with a \$9 billion backlog of maintenance needs. \\
    (intrinsic) generated & \hlarg{America's 391 national parks are facing} a \$9 billion backlog of maintenance needs. \\
\bottomrule          
\end{tabular}
\caption{\falsesum{}-generated summaries that are unintentionally consistent with the source document. \hlarg{Green-highlighted} spans indicate information which is consistent with the document.}
\label{tab:fail_cases}
\end{table*}

\begin{table*}
\centering
\footnotesize
\begin{tabular}{rl}
\toprule
\toprule
\multicolumn{2}{p{16cm}}{
    \predbox{$\texttt{Predicates}$}: \hlpred{is being offer for, were steal from, sell, Both as a solo artist and leader of the Heartbreakers, is one of , according to, where were rehearse for, contribute to, was induct into in}; \argbox{$\texttt{Arguments}$}: \hlarg{the Heartbreakers, The band, Denise Quan, five guitars, the Recording Industry Association of America, more than 57 million albums, Petty, A 7,500 reward, a soundstage, the Rock \& Roll Hall of Fame}; \codebox{$\texttt{Code}$}: \hlcode{intrinsic}; \sumbox{$\texttt{Summary}$}:\hlblank{1} \hlblank{0} \hlsum{the 1960s.}
}\\
\midrule
    gold & Three of them were vintage guitars from the 1960s. \\
    (intrinsic) generated & \hlerror{The band was inducted into the Rock \& Roll Hall of Fame in} the 1960s. \\
\midrule
    \multicolumn{2}{p{16cm}}{
    \predbox{$\texttt{Predicates}$}: \hlpred{: is only the second time in, How could have do with, was lace with, struggle against at, have score, expect to match, had settle into, ignite, has lost, Just as was walk into, were already circulate on, begin to filter, watch on in}; \argbox{$\texttt{Arguments}$}: \hlarg{his chair, Anfield, clips, the stands, symbolism, 13 Premier League goals, Brendan Rodgers, through, Liverpool, the 100-plus strikes of last season, 13 games against Hull, everything, one}; \codebox{$\texttt{Code}$}: \hlcode{intrinsic}; \sumbox{$\texttt{Summary}$}:\hlsum{Luis Suarez took three minutes to} \hlblank{0} \hlblank{1}.
}\\
\midrule
    gold & Luis Suarez took three minutes to get his first assist for Barcelona. \\
    (intrinsic) generated & Luis Suarez took three minutes to \hlerror{ignite symbolism}. \\
\midrule
    \multicolumn{2}{p{16cm}}{
    \predbox{$\texttt{Predicates}$}: \hlpred{allegedly know, supposedly write, in ' was underway, is investigate, file against in by, file in, forbid, was toss by in, wait for, fire at, accuse of, decide to fire based on, new information state, told, allegedly sent to, was complicate by, Even though was toss, allegedly made, hold no more, expose to}; \argbox{$\texttt{Arguments}$}: \hlarg{the case, new information states, his sexual abuse, more recent damages, people, the blog posts, 2011, him, This week, her, allowing at one of his Los Angeles stores to post naked photos of Morales on a blog that was meant to appear as though it belonged to Morales, American Apparel, The Post, a settlement, The clothing company, Charney, new information saying he allowed an employee to impersonate and post naked photos online of an alleged victim of his sexual abuse who filed a case against him in 2011, a settlement 'in the low six-digits' was underway, the company title, employee, 2012, The \$260 million lawsuit, a report from March 25, 2011 that said Morales allegedly sent nude photos of herself to Charney after she stopped working at the store, nude photos of herself, Morales}; \codebox{$\texttt{Code}$}: \hlcode{extrinsic}; \sumbox{$\texttt{Summary}$}:\hlsum{Women in the video} \hlblank{0} \hlblank{1}.
}\\
\midrule
    gold & Women in the video have been identified as current or former American Apparel workers. \\
    (extrinsic) generated & Women in the video \hlerror{were allegedly sexually assaulted by Morales}. \\
\bottomrule          
\end{tabular}
\caption{Examples of the formatted input at test time and the real output of the \falsesum{} generation model. \hlpred{Blue-highlighted} spans show the formatted input predicates. \hlarg{Green-highlighted} spans show the formatted input arguments. \hlcode{Yellow-highlighted} spans show the formatted input control code. \hlsum{Gray-highlighted} spans show the formatted input masked gold summary. \hlerror{Red-highlighted} spans show the information inserted by the model to render inconsistent summaries.}
\label{tab:input_output_examples}
\end{table*}